\documentclass{article}
\usepackage{arxiv}
\usepackage[utf8]{inputenc} 
\usepackage[T1]{fontenc}    
\usepackage{hyperref}       
\usepackage{url}            
\usepackage{booktabs}       
\usepackage{amsfonts}       
\usepackage{nicefrac}       
\usepackage{microtype}      
\usepackage{lipsum}		
\usepackage{graphicx}
\usepackage{natbib}
\usepackage{doi}
\usepackage{float}

\title{On the global feature importance for interpretable and trustworthy heat demand forecasting}

\date{} 					

\author{ 
    \href{https://orcid.org/0000-0002-4141-2058}
        {\includegraphics[scale=0.06]{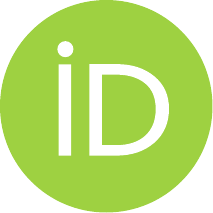}\hspace{1mm}Milan Zdravković}
            \thanks{This research was supported by the Science Fund of the Republic of Serbia, Grant No. 23-SSF-PRISMA-206, Explainable AI-assisted operations in district heating systems - XAI4HEAT} \\
	Faculty of Mechanical Engineering\\
	University of Niš\\
	ul. Aleksandra Medvedeva 14, 18000 Niš, Serbia \\
	\texttt{milan.zdravkovic@masfak.ni.ac.rs} \\
}

\renewcommand{\headeright}{}
\renewcommand{\undertitle}{}

\hypersetup{
pdftitle={A template for the arxiv style},
pdfsubject={q-bio.NC, q-bio.QM},
pdfauthor={David S.~Hippocampus, Elias D.~Striatum},
pdfkeywords={First keyword, Second keyword, More},
}

\begin{document}
\maketitle
\begin{abstract}
	The paper introduces the ante-hoc Explainable AI methodology to assess the global feature importance of the Machine Learning models used for heat demand forecasting in intelligent control of District Heating Systems, with motivation to facilitate their interpretability and trustworthiness, hence addressing the challenges related to adherence to communal standards, customer satisfaction and liability risks. Methodology includes use of four different approaches, namely intrinsic interpretability of Gradient Boosting method and selected post-hoc methods, namely Partial Dependence, Accumulated Local Effects and SHAP. None of the selected methods assume feature permutation or perturbations which can introduce bias due to introduction of random unrealistic values of data instances. Discussion of results is provided, including the assessment of complementarities where applicable, with specific interpretations in context of the district heating processes.
\end{abstract}

\bigskip
\noindent\textbf{Publication note.}
This preprint corresponds to the paper published as:

Zdravkovic, M. (2025). On the global feature importance for interpretable and trustworthy heat demand forecasting. Thermal Science, 00, 48–48.  
The final authenticated version is available at
\url{https://doi.org/10.2298/TSCI241223048Z}.

\section{Introduction}
Machine Learning methods have already proven useful in enabling intelligent control of District Heating Systems (DHS), for example by replacing the trivial automatic control mechanisms based on a control curve and affected only by the ambient temperature, with proactive, intelligent control approach driven by complex multi-variate heat demand forecasting methods \cite{Zdravkovic2022ExplainableHeat}\cite{Runge2023Comparison}\cite{Wei2021Prediction}.

Given its hourly seasonality and predictable control patterns, the heat demand forecasting is not considered a difficult problem. It is being addressed by using time series forecasting algorithms, such as Long-Short Term Memory (LSTM) \cite{Hochreiter1997LSTM}, but also with the conventional ML algorithms, such as highly effective gradient boosting \cite{Gong2020GBM}\cite{Zdravkovic2024XGBoost}, after the transposition of training data to classical tabular format. Still, importance of the role of heat demand forecasting step in overall control strategy poses quite a few potential risks, including those related to compliance (for example, adherence to CO2 emission levels), customer satisfaction, and in some cases even liability due to incapacity to meet the contractually agreed service levels (for example, consumers room temperature). The key challenge here is that time series forecasting algorithms are in charge of making very important decisions which are not so easy (if not impossible) to justify due to their intrinsic complexity, which cannot be perceived and understood by the stakeholders in heat delivery process, including domain experts and plant operators. The motivation behind the research presented in this paper is to discuss possible insights into interpretability of those decisions and models that make them. In order to do so, an Explainable AI (XAI) approach will be used.
Explainable AI is a young scientific discipline \cite{Ribeiro2016LIME} which refers to set of concepts and methods aiming at facilitating transparency (models are understandable), interpretability (decisions can be justified), accountability (why the decision has been made, because of model architecture, hyperparameter setting, data, etc.), fairness and bias detection (are models discriminating against specific features), and trust in AI models. XAI can be considered on two different levels of interpretation: ante-hoc (before the fact) and post-hoc (after the fact) interpretation. Ante-hoc refers to the models which are inherently interpretable, such as linear regression or decision trees, so there is no need for additional tools. Post-hoc refers to methods to interpret and explain the model (typically considered as black box) after it was trained and deployed.

XAI methods can be classified as model-agnostic or model-specific. Model-agnostic methods are those which can be applied to any AI model, regardless of its structure. Common approach to model-agnostic methods is the use of so-called surrogate models, where inherently explainable models (such as linear regression or decision trees) are used to approximate (either locally or globally) complex, not explainable models, such as deep neural networks. Model specific methods are those which use some features that have contribution to explainability but are intrinsic to the model, for example, attention scores in Transformer networks or decision trees in gradient boosting models. Sometimes, those methods are classified as black-box (model-agnostic) and white-box (model-specific) methods. Finally, methods can be classified based on their scope. Local methods are those which explain an individual prediction. Global methods interpret entire model behavior.

Most usually, XAI is used to achieve a human understanding of how very complex ML algorithms make decisions by providing justifications in the form of interpretable insights. Another important aspect of XAI is that many of its global insights can provide a human readable argument for checking validness of the model by the experts. This is very important for trustworthy automation of impactful systems, such as DHS, especially when considering out-of-distribution incoming data for forecasts.
In the continuation of this paper, the methodology for addressing the global feature importances of the heat demand forecasting model is described. The methodology includes the basic description of Machine Learning pipeline to train the model and introduces specific approaches to assessing the feature importances. In the third section, the results of those approaches when applied on the trained model are presented and discussed. Finally, in the fourth section, the key conclusions are highlighted.

\section{Methodology}
\label{sec:headings}

The research presented in this paper aims at identifying the effective approaches to evaluating trustworthiness of the heat demand forecasting models by using different aspects of interpretability of feature importances at the global level, for the whole trained model. The test model to interpret is trained by using ensembles of decision trees. Such an approach is chosen so to be able to compare and combine different metrics of feature importances exposed intrinsically by the model and those delivered by the post-hoc interpretability methods. The model which will be evaluated is regression model which makes hourly forecasts of the transmitted heat energy by using 4 heating seasons of historical data. The model is trained on the pre-processed data within the pipeline described in \cite{Zdravkovic2024XGBoost}.

Historical data from SCADA (Supervisory Control and Data Acquisition) system of substation 17 of local DHS, managed by the Faculty of Mechanical Engineering, will be used in this demonstration. The data encompasses 4 heating seasons (2020-2024) and is merged with meteorological data to introduce additional predictors. Figure 1 illustrates the time series data for the selected features in the given interval. Data is acquired from SCADA system, and it includes hourly transmitted heating energy (deltae), ambient temperature ($t_{amb}$), water temperature in the secondary supply line (C, $t_{sup_sec}$), water temperature in the secondary return line (C, $t_{ret_sec}$), water temperature in the primary supply line (C, $t_{sup_prim}$) and water temperature in the primary return line (C, $t_{ret_prim}$). Besides this, various other hourly meteorological data is included in the dataset, such as feels-like temperature, dew, wind direction and others.

Preprocessing activities include stripping all data points except those acquired at full hour, inserting missing timepoints and populating those with missing values (NaN), treatment of zero data in the column of energy readings by replacing those values with missing values, introducing basic time features such as time of a day, replacing all missing values by using linear interpolation, removing data outside of the heating season, removing all features with Pearson correlation coefficient less than 0.1 to reduce dimensionality of the model, replacing data points with relatively small values of transmitted energy close to zero - with zero (corresponding to the inaccurate readings), and removing outliers (z-score>4) and their replacement with linearly interpolated values. All those tasks are elaborated and justified in more detail in \cite{Zdravkovic2024XGBoost}. Then, the processed data is used to train the regressor (by using XGBoost algorithm) model. Before training, data is enriched with time-lagged transmitted energy (deltae-1 to deltae-23) and ambient temperature (temp-1 to temp-23) in the past 24 hours which are introduced as new features. 80\% of available data is used for training and the remaining 20\% is used for testing the trained models. The output shows Mean Absolute Error (MAE) as adopted simplistic metric for regression problem. The model was not optimized or associated with more advanced metrics because the purpose of this research is to investigate the opportunities of use of XAI insights.

\begin{figure}[H]
	\centering
	\includegraphics[width=0.75\textwidth]{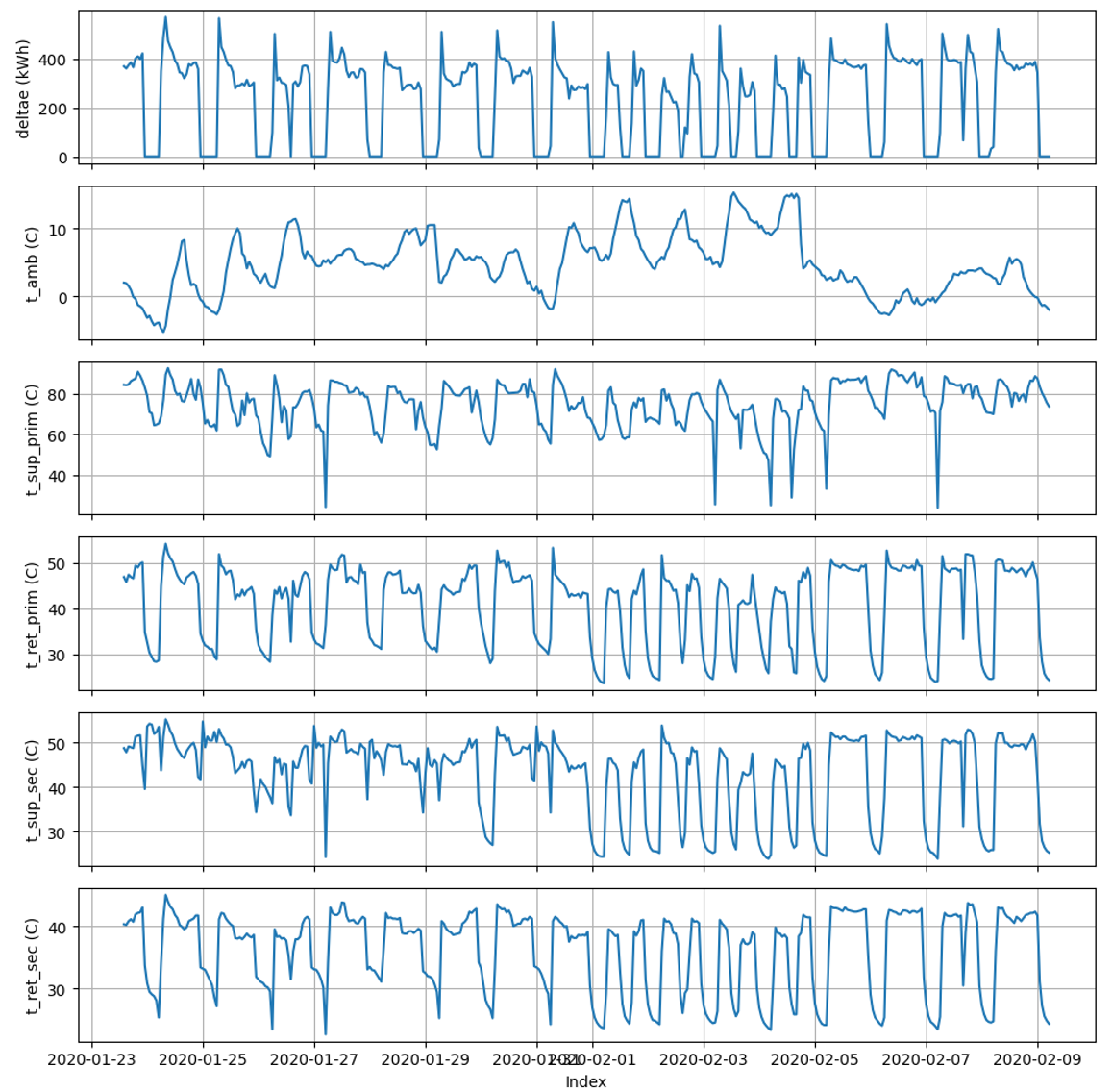}
	\caption{Line plots of the selected data features in example interval.}
	\label{fig:fig1}
\end{figure}

Global feature importance will be discussed by using intrinsic interpretability of gradient boosting method and selected ante-hoc methods, namely Partial Dependence, Accumulated Local Effects and SHAP. Partial Dependence (PD) \cite{Friedman2001GBM} is the measure of how one model prediction varies with respect to a feature of interest. PD plots show how the target variable changes over the distribution of the selected input variable - a feature of interest. Main limitation of PD is the assumption of feature independence. Natural extension of PD are Accumulated Local Effects (ALE) plots \cite{Apley2020ALE}. Unlike PD, it doesn't struggle with dependencies in the underlying features. SHAP (SHapley Additive exPlanations) \cite{Lundberg2017SHAP} values come from cooperative game theory and are used to fairly distribute the total gain (or loss) among players based on their contributions. In the context of Machine Learning, the "players" are the features of the model, and the "gain" is the prediction made by the model. A Shapley value for a feature represents its average contribution to the prediction across all possible combinations of features. SHAP can be used for both local and global explainability. SHAP values can explain individual predictions by showing how each feature contributes to that specific prediction. By aggregating SHAP values across many instances, SHAP can also provide insights into the overall importance of features and how they generally influence the model's predictions.

\section{Overview of global feature importance insights and discussion}

In this section, comprehensive set of approaches to interpret different aspects of the feature importance in the trained models for forecasting heat demand will be demonstrated, namely gain/cover/frequency aspects of the global feature importances, Partial Dependence, Accumulated Local Effects and SHAP.

\subsection{Gain, cover and frequency metrics of algorithms based on Decision Trees}

Decision trees (DT) and linear regression (LR) are inherently explainable algorithms. While LR is based on the weights associated to each of the features in the dataset whose scale indicates its importance, decision trees infer decision rules from the data features. DT works by recursively splitting the dataset into subsets based on the feature that provides the most significant information gain or reduces impurity the most. This process creates a tree-like model of decisions, where each internal node represents a feature-based decision rule, each branch represents the outcome of that decision, and each leaf node represents a final prediction or outcome.

The Gradient Boosting algorithm is so-called Machine Learning ensemble method, which comprises of multiple decision trees built sequentially, and which actually predict residuals, namely differences between previously predicted values and actual values (for regression problems). In the first step, initial prediction is normally mean value of the target variable, and the first decision tree is trained to predict the differences between actual value and this, mean value. Then, predicted residuals from the previous step are again added to the initial prediction, new predicted value is calculated and in the following step, residuals are used to train third decision tree. The process can be described with:

\begin{equation}
\hat{y}_i^{(M)} = F_M(X_i)
= F_0 + \sum_{m=1}^{M} f_m(X_i)
\end{equation}

\begin{itemize}
    \item $\hat{y}_i^{(M)}$ is the predicted value for the $i$-th data point after $M$ iterations.
    
    \item $F_M(X_i)$ is the model's prediction after $M$ iterations.
    
    \item $F_0$ is the initial prediction (base score). For regression,
    $F_0 = {mean}(y)$.
    
    \item $f_m(X_i)$ is the prediction of the $m$-th tree (namely, the residual predicted by the $m$-th tree for the $i$-th data point).
\end{itemize}

Gradient Boosting algorithms are inherently explainable as they provide out-of-the-box analysis of gain, cover and frequency, which are all different aspects of the feature importance. Gain measures the improvement in accuracy (or reduction in loss) that a feature provides when it is used in a split. Gain is a direct measure of a feature's contribution to the accuracy of the model. If a feature consistently results in large gains when used in splits, it means that this feature is very effective in making accurate predictions. Cover measures the relative quantity of observations (or data points) that are affected by a feature when it is used in a split. Higher cover indicates that the feature is used in splits that affect a large portion of the dataset. If a feature has high cover, it means that it was consistently found as significant for all or large number of predictions. Frequency (or weight) measures the number of times a feature is used in all the trees of the model. Higher frequency indicates that the feature is frequently used to make splits in the model. It shows how often the feature is chosen to make a decision.

Figure 2 shows frequency, gain and cover analysis of the feature importances of Gradient Boosting model for heat demand forecasting.

\begin{figure}[H]
	\centering
	\includegraphics[width=1\textwidth]{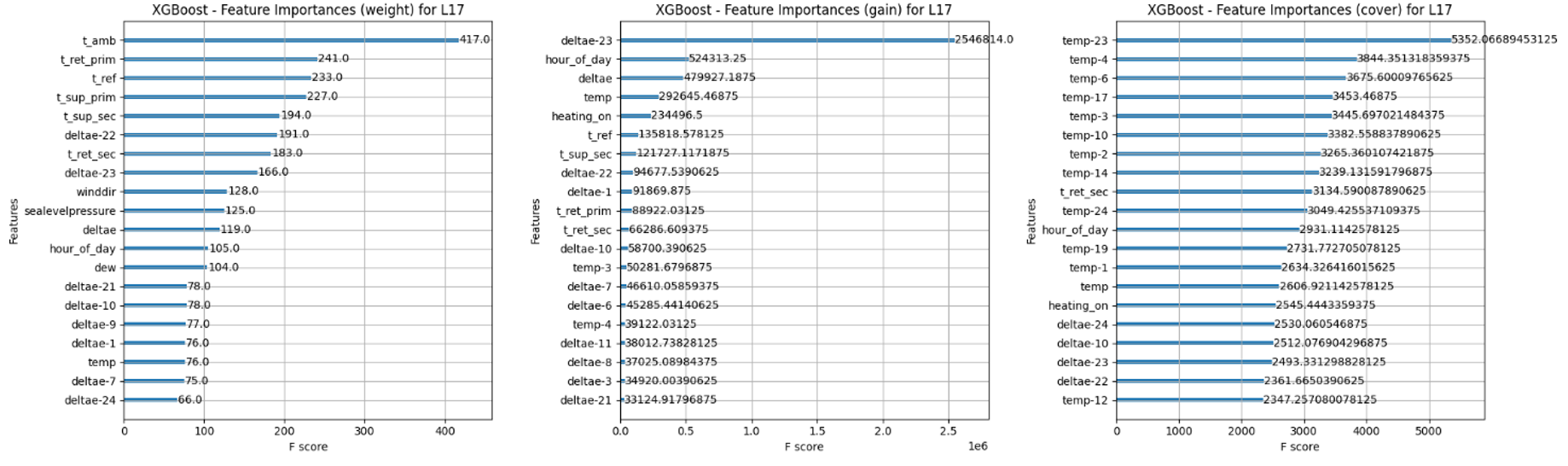}
	\caption{Frequency, gain and cover analysis of the feature importances.}
	\label{fig:fig2}
\end{figure}

Weight bar plot illustrates that ambient temperature ($t_{amb}$) is the most used feature in decision making across multiple trees in Gradient Boosting algorithm, composed of 100 boosted trees, indicating that it is considered to be an important decision point. Transmitted energy at the same hour of the previous day (deltae-23) is by far the most influential in improving the model's performance when used in splits, indicating strong daily seasonality of DHS operation. Finally, temperature reading by the sensor placed near to the substation in the same hour of the previous day (temp-23) has the larger cover, meaning that it was relevant decision point for most of the instances in the test dataset. All those interpretations directly correspond to common sense and a real-life situation, meaning that the trained model accurately reflects the complex dependencies within DHS, making it more trustworthy and reliable.

\subsection{Partial Dependence}

Partial Dependence is a post-hoc XAI technique to assess the feature importance to the predicted outcome, while ignoring the effects of other features. PD plots provide visual interpretations of how a feature influences the prediction across different values, helping to understand the model's behavior globally. Flat PD plot implies little or no effect of one to another variable, meaning that specific quantified measure of one feature importance is a variance of the PD values across the distribution of that feature.
Individual Conditional Expectation (ICE) plots provide a more granular view by showing how the prediction changes for individual instances as a particular feature varies. ICE plots allow for the detection of heterogeneous effects and feature interactions that might be averaged out in PD assessment. Again, it’s important to highlight that Partial Dependence captures only the main effect of the feature and ignores possible feature interactions. In other words, when there are mutual correlations between two features, PDP cannot be trusted.

ICE is plotted for four selected features (see Figure 3), namely transmitted energy in the same hour of the previous day (deltae-23), hour of the day, temperature measured by DHS (t\_amb) and the one measured by the nearby meteorological station (temp). The average line (orange) represents PD. The y-axis of a Partial Dependence Plot (PDP) represents the partial dependence of the target variable (average predicted outcome) on the feature(s) being plotted, while marginalizing over the other features in the model. The vertical ticks at the bottom of each plot represent the distribution of the data points in the dataset. The density of ticks suggests where most data points are located for each feature.

\begin{figure}[H]
	\centering
	\includegraphics[width=1\textwidth]{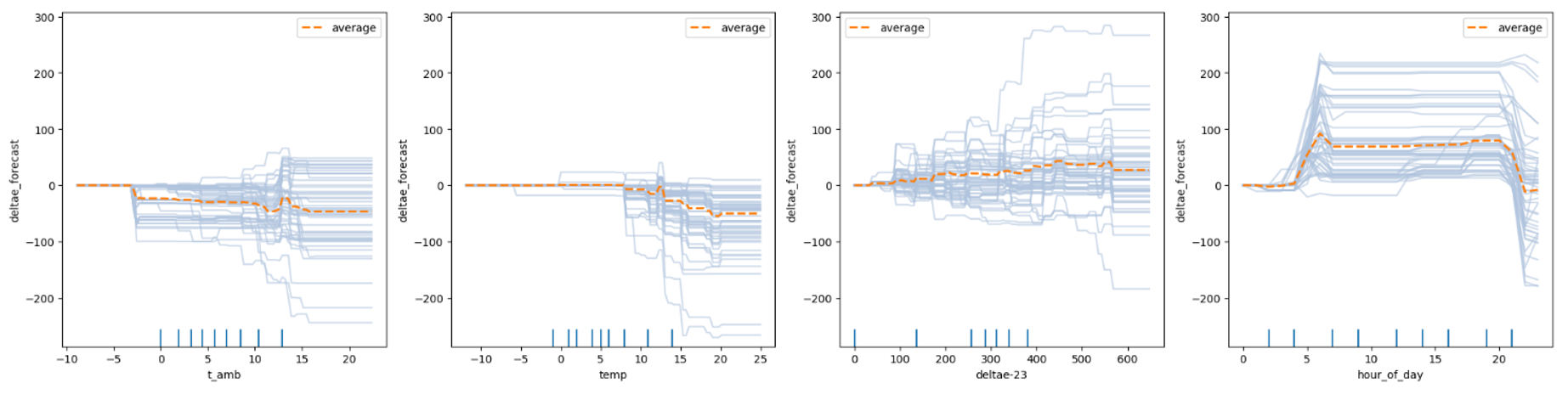}
	\caption{ICE plots for the selected features in the dataset.}
	\label{fig:fig3}
\end{figure}

PD line indicates that higher values of $t_{amb}$ are generally associated with a decrease in the predicted value, but this trend reverses at the specific point and shows the spike at approximately $t_{amb}$=13, which could be indicating some non-linear effects in the model or feature interactions. The PD plot of deltae-23 is mostly flat, especially for delta-23<300. The PD plot of hour\_of\_day follows the change pattern from the time series, confirming the repeatability of this pattern and strongly hourly seasonality. Predictions are significantly higher during certain hours of the day, particularly in the early morning to late afternoon, but drop sharply in the late evening and early night hours. This feature captures a time-related pattern and strongly influences the model's predictions.

Under the assumption of lack of feature interaction, characteristic points in PD plots can be understood as decision points of the XGBoost model. We can take on this to interpret the lack of responsivity of the target variable on the changes of temperature measured at the meteorological station for temp<7 (second plot). Weight metrics (see Figure 2) show that $t_{amb}$ (temperature measured in the substation) is used as a decision point significantly more frequently than temp, even though those two signals are very similar.

Also, flatness of deltae-23 PD plot could imply its non-existing effect on target variable, despite a very high weight calculated by XGBoost feature importance method, implying a contradiction. However, it must be taken into account that PD plot provides a global view of how a feature influences the model's predictions by averaging the model's predictions over the marginal distribution of the feature. In case of deltae-23, ICE plots show that average is not representative of the contributions over individual instances having quite diverse plots.

\subsection{Accumulated Local Effects}

The method of Accumulated Local Effects (ALE) exhibits much less sensitivity to the mutual correlations of the input features, namely their interactions and thus, it is considered as much more reliable method for determining the feature importance, in general. Instead of assuming feature independence, ALE calculates the local effects of a feature by looking at how small changes in the feature value affect the prediction within the observed data distribution. Figure 4 shows ALE plots for selected four features.

The distribution of ALE for temperatures is more stable than in case of PDP. It shows that transmitted energy will decrease almost linearly in the interval of $t_{amb}$<9. Then, the decrease rate will become more rapid with some non-linearities. Flatness of ALE plot for deltae-23 feature is already previously explained.

\begin{figure}[H]
	\centering
	\includegraphics[width=1\textwidth]{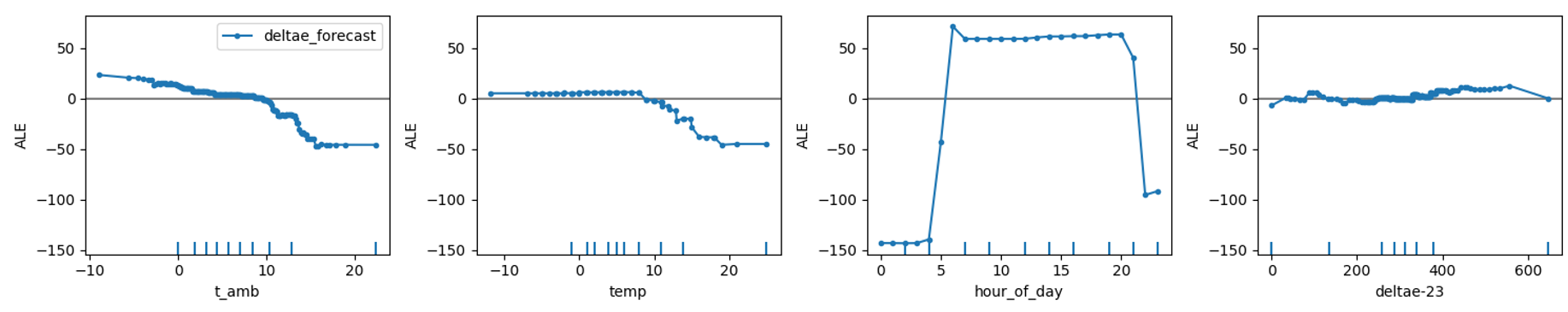}
	\caption{ALE plots for the selected features in the dataset.}
	\label{fig:fig4}
\end{figure}

\subsection{SHAP approach to assessing global feature importance}

SHAP and LIME (Local Interpretable Model-agnostic Explanations) \cite{Ribeiro2016LIME} are two most commonly used methods for assessing feature importances. Both methods are model agnostic, with one major difference in scope, as LIME cannot be used as is for interpreting global feature importance. Even though SHAP is more computationally expensive, it is considered as more reliable, even in local interpretability, due to its strong theoretical foundation (cooperative game theory) which becomes important especially in more complex models with high uptake of non-linearity (LIME's linear approximation might not always capture the true local behavior of the model).

For each feature, SHAP calculates the contribution by considering the difference in the model's prediction with and without the feature - a marginal contribution. This is done by evaluating all possible combinations of feature subsets (also known as coalitions) with and without the feature and averaging the marginal contributions of the feature across all these combinations. The SHAP value for a feature i in a specific prediction is computed as:

\begin{equation}
\phi_i
=
\sum_{S \subseteq N \setminus \{i\}}
\frac{|S|!\left(|N|-|S|-1\right)!}{|N|!}
\left[f(S \cup \{i\}) - f(S)\right]
\end{equation}

\begin{itemize}
    \item $S$ is a subset of features not including feature $i$.
    \item $N$ is the set of all features.
    \item $|S|$ is the number of features in subset $S$.
    \item $f(S)$ is the model's prediction given the features in subset $S$.
    \item $f(S \cup \{i\})$ is the model's prediction given the features in subset $S$ plus feature $i$.
    \item
    $\frac{|S|!\left(|N|-|S|-1\right)!}{|N|!}$ is the weighting factor that ensures fair distribution of the marginal contribution across all possible subsets.
    \item $\phi_i$ is the SHAP value for feature $i$, representing its contribution to the difference between the actual prediction and the baseline prediction.
\end{itemize}

Global SHAP feature importance is determined by averaging the absolute SHAP values for each feature across all instances in the dataset.

\begin{figure}[H]
	\centering
	\includegraphics[width=1\textwidth]{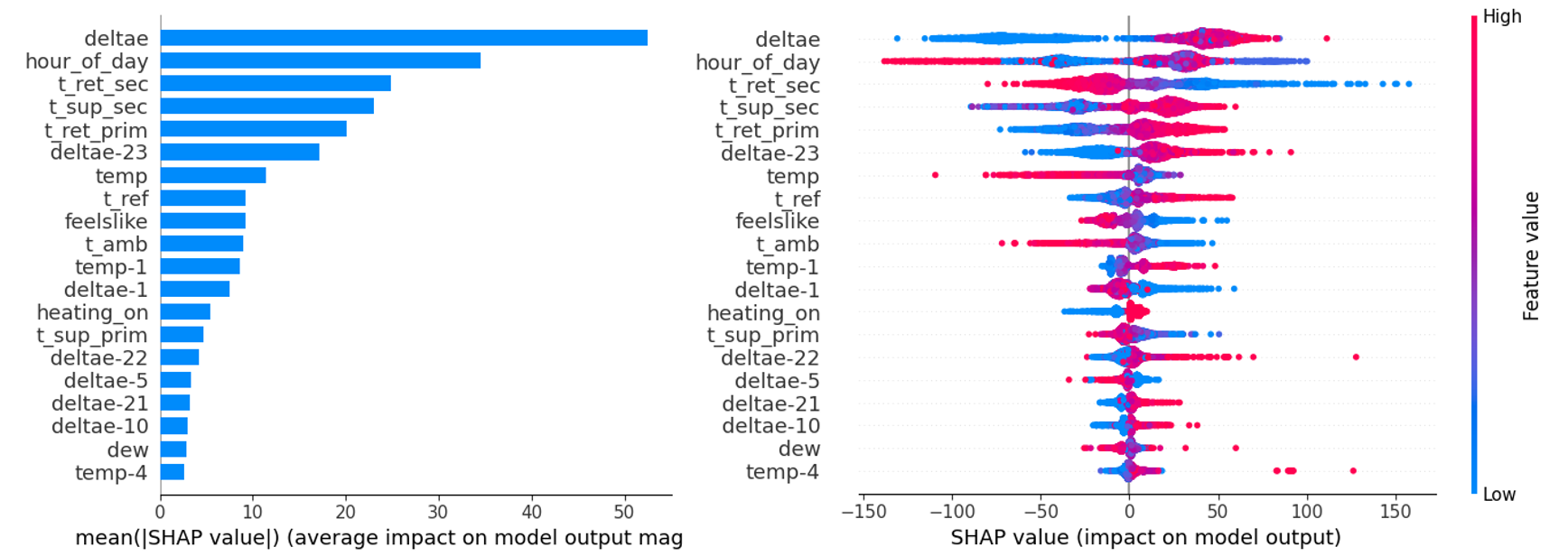}
	\caption{ALE plots for the selected features in the dataset.}
	\label{fig:fig5}
\end{figure}

Figure 5 shows the bar plot with ranked SHAP feature importances (on the left) and SHAP summary plot (on the right) for the heat demand forecasting model.

SHAP summary plot provides a more comprehensive global feature importances view. Each row in the plot represents a feature and the features are ordered by their importance. Positive SHAP values (to the right) increase the value of model's prediction, while negative SHAP values (to the left) decrease the prediction. The density of points at each SHAP value shows the distribution of the SHAP values for each feature. The wider the plot at a certain SHAP value, the more data points have that SHAP value.

Features with the widest range of SHAP values, namely deltae (transmitted energy at the current timepoint) and hour\_of\_day has the greatest impact on the model's output. Based on the color gradient, it can be concluded that high values of deltae increase the forecast (SHAP values > 0), while low values (SHAP values < 0) decrease it. It’s clearly the opposite case for $t_{ret_sec}$ and feelslike, for example. This all clearly corresponds to how DHS work.

It is important to clarify notable differences with the results of gain, cover and weight metrics analysis, presented by the XGBoost feature importance function. First, it is important to highlight that intrinsic explainability features of the model are rooted in one set of data - training dataset, while post hoc interpretability feature importances are calculated based on another - a test dataset. Furthermore, weight and cover as measures of frequency of feature use in a split and the number of instances affected by the split, respectively, are complementary to the SHAP importances and cannot be compared with it. A feature might have high weight or cover, but low SHAP values if it is used frequently or broadly but doesn't significantly affect the model's predictions. However, comparing gain with SHAP values can validate whether the features that improve model accuracy the most are also the ones that most influence predictions. If a feature has high SHAP values but low gain, it might suggest that the feature is important for predictions but doesn't improve model accuracy as much during training. This could happen if the feature captures information similar to other features (redundancy). In DHS forecasting model, this is the case with $t_{amb}$ - it has moderately high SHAP values, but it is not even ranked in the gain bar plot. In the opposite direction, a feature with high gain but low SHAP values might be crucial for certain splits but doesn't have a large overall impact on predictions, possibly because its effect is overshadowed by other features or only relevant in specific scenarios. This is the case with deltae-23, since the repeatability of DHS operation patterns is highly distinguished at the beginning of the workday (when the heating starts and in first few hours) but less towards its end when operator starts changing the control curve based on the ambient temperature.

\section{Conclusions}

This paper introduces global feature importance calculation methods with intended use to validate the trained forecasting model in environments requiring legal compliance and agreed service levels on a massive customer scale, which is all the case with District Heating services. The provided discussion elaborates on the specific interpretations of different XAI ante-hoc and post-hoc methods in context of DHS’s expected behavior and its human, domain expert interpretation. Four methods to elaborate the global feature importance are implemented in case of the model for forecasting heat demand, trained by using Gradient Boosting ML algorithm, namely, ante-hoc feature importances, Partial Dependence, Accumulated Local Effects and SHAP. Some widely used, well-known techniques have been discarded from the beginning because of assuming feature permutation or perturbations which can introduce bias due to introduction of random unrealistic (out of realistic distribution) values of data instances and linear dependencies. This is the case with Permutation Importance \cite{Fisher2019ModelClassReliance} and LIME. It is important however to highlight that both PD and ALE also use marginal distributions of input variables but only with minor variances, when compared to above mentioned two methods.

However, Partial Dependence is the simplistic method which captures only the main effect of the feature and ignores possible feature interactions. In other words, when there are mutual correlations between two features, PD plots are not a reliable source for interpretation. Still, ICE plots allow for the detection of heterogeneous effects and feature interactions that might be averaged out in PD assessment. Heterogeneous effects are here spotted for deltae-23 feature; transmitted energy at the same hour of the previous day is a good predictor of heat demand in the next hour, but mostly for the opening hours of daily operation. ALE has been proven as more reliable method for determining the feature importance, taking into account feature interactions and nonlinearities. 

ALE plots clearly reveal that transmitted energy will decrease almost linearly in the interval of $t_{amb}$<9; For warmer temperatures, the decrease rate will become more rapid with some non-linearities. This behavior closely resembles the shape of the control curve, which determines the dependence of water temperature in secondary supply flow ($t_{sup_sec}$) with ambient temperature ($t_{amb}$); the resemblance is actually the strong evidence of quality of forecasting model.

SHAP’s assessment of global feature importances is based on a strong foundation provided by the game theory and it accounts for feature interactions while also addressing the nonlinearities in the model, which all makes it the most reliable and robust method. This is proven in the case of interpreting the feature importances of the heat demand forecasting model.

Benefits of using XAI techniques and specifically interpretation of global feature importances in heat demand forecasting models are numerous. Transparency of otherwise black-box models helps build trust in the model’s predictions and decisions driven by those predictions. Identifying key features can guide operational strategies, such as adjusting heating schedules or optimizing energy use based on predicted demand. Efficiency improvements become more obvious: by focusing on important features, operators can make targeted improvements to system efficiency, potentially reducing energy consumption and costs. Insights from feature importance can support long-term planning and strategy development for infrastructure investment and system upgrades. Understanding which features are critical can help in refining the model, addressing potential weaknesses, and improving overall accuracy.

\bibliographystyle{unsrt}
\bibliography{references}

\end{document}